\documentclass[letterpaper]{article} 
\usepackage{aaai2026}  
\usepackage{times}  
\usepackage{helvet}  
\usepackage{courier}  
\usepackage[hyphens]{url}  
\usepackage{graphicx} 
\usepackage{natbib}  
\usepackage{caption} 
\usepackage{makecell}
\usepackage[linesnumbered,ruled,vlined]{algorithm2e}
\usepackage{amsmath}
\usepackage{subcaption}
\usepackage{booktabs}
\usepackage[table]{xcolor}
\usepackage{tikz}
\usetikzlibrary{shapes.geometric, arrows, positioning}

\usepackage{newfloat}
\usepackage{listings}
\DeclareCaptionStyle{ruled}{labelfont=normalfont,labelsep=colon,strut=off} 
\title{TCIA: A Task-Centric Instruction Augmentation Method\\for Instruction Finetuning}
\author {
    Simin Ma \textsuperscript{\rm 1,*,\dag}, Shujian Liu \textsuperscript{\rm 1,*}, Jun Tan \textsuperscript{\rm 1,*}, Yebowen Hu \textsuperscript{\rm 1}, Song Wang \textsuperscript{\rm 1}, Sathish Reddy Indurthi \textsuperscript{\rm 1}, Sanqiang Zhao \textsuperscript{\rm 1}, Liwei Wu \textsuperscript{\rm 1}, Jianbing Han \textsuperscript{\rm 1}, Kaiqiang Song \textsuperscript{\rm 1}
}
\affiliations {
    \textsuperscript{\rm 1}Zoom Communications Inc.\\
    \textsuperscript{*}Equal contribution.\\
    \textsuperscript{\dag}Corresponding author: simin.ma@zoom.us
}

\usepackage{bibentry}

\begin{document}

\maketitle

\begin{abstract}
Diverse instruction data is vital for effective instruction tuning of large language models, as it enables the model to generalize across different types of inputs . Building such diversified instruction dataset is an essential step in this process. Existing approaches often leverage large language models to automatically explore and generate diverse instructions, ensuring both data diversity and quality. However, they tend to overlook an important factor in real-world applications: on-task relevance. In practice, only a few real-world applications require a truly general-purpose model; most benefit from task-specific knowledge tailored to their particular use case. Therefore, it is vital to develop instruction augmentation methods that not only maintain diversity but are also optimized for specific, real-world scenarios.

We thus introduce \textbf{T}ask \textbf{C}entric \textbf{I}nstruction \textbf{A}ugmentation (TCIA), a framework that systematically expands instructions while preserving both diversity and task alignment. By representing instructions in a discrete query-constraints space, TCIA creates a rich set of task-relevant instructions and enables models to generalize to these task-specific instructions without sacrificing overall performance. Experiments show that TCIA improves open-source LLMs' performance by an average of 8.7\% across four real-world, task-specific applications, and in some cases outperforming leading closed-source models. These improvements do not compromise general instruction-following ability, making TCIA a scalable and efficient solution for adapting LLMs to real-world, task-focused applications.
\end{abstract}

\section{Introduction}
\begin{figure}[ht]
\centering
\begin{subfigure}{0.22\textwidth}
    \centering
    \includegraphics[width=\linewidth]{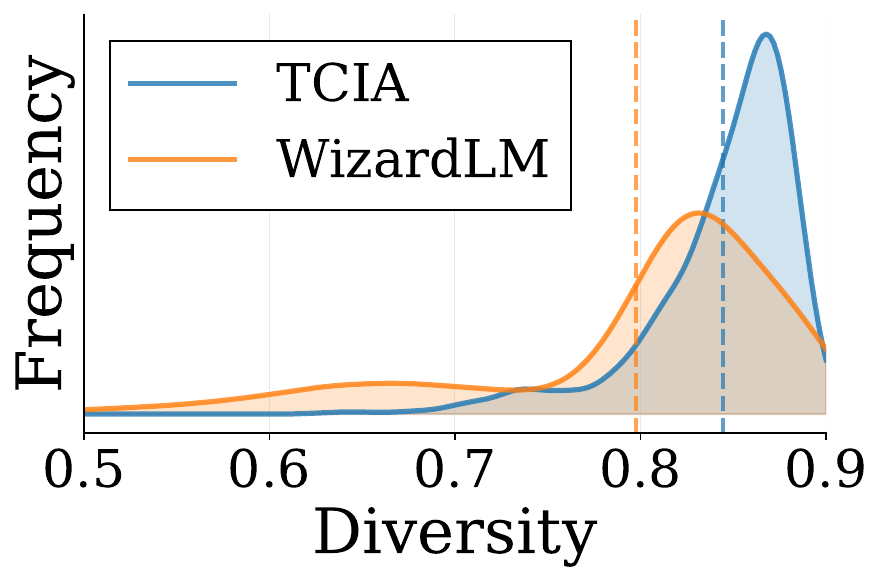}
    \caption{\centering Hop 1}
    \label{fig: diversity_density_hop1}
\end{subfigure}
\hfill
\begin{subfigure}{0.22\textwidth}
    \centering
    \includegraphics[width=\linewidth]{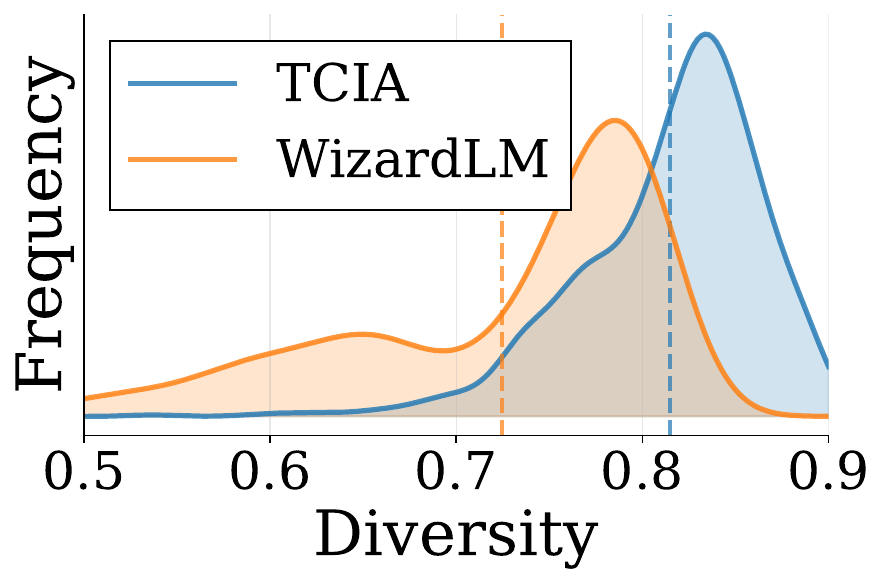}
    \caption{\centering Hop 2}
    \label{fig: diversity_density_hop2}
\end{subfigure}

\vspace{0.5em}

\begin{subfigure}{0.22\textwidth}
    \centering
    \includegraphics[width=\linewidth]{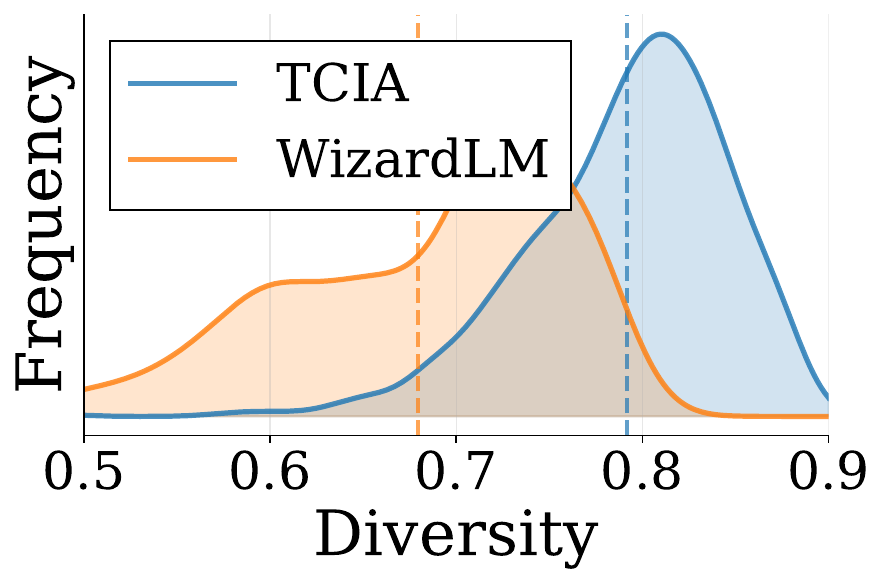}
    \caption{\centering Hop 3}
    \label{fig: diversity_density_hop3}
\end{subfigure}
\hfill
\begin{subfigure}{0.22\textwidth}
    \centering
    \includegraphics[width=\linewidth]{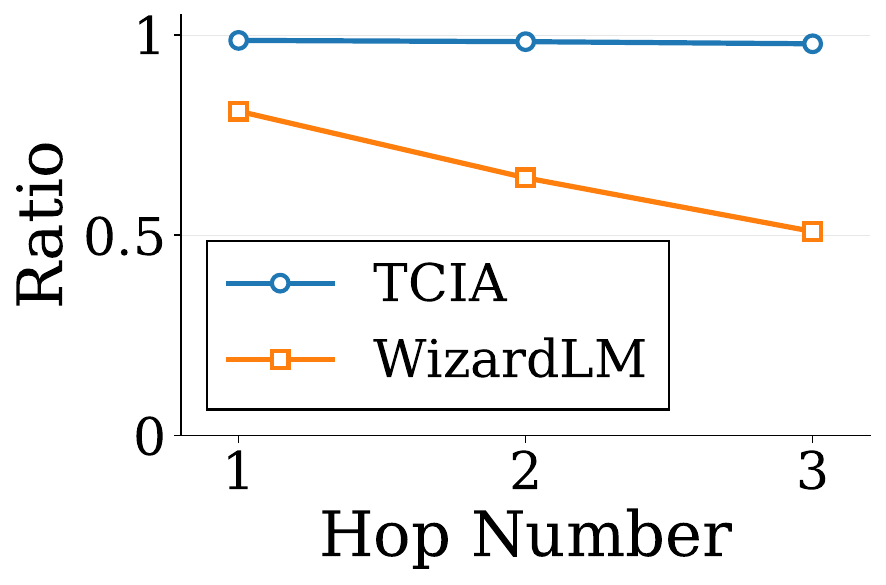}
    \caption{\centering On-Task Ratios}
    \label{fig: ontask_average}
\end{subfigure}
\caption{Comparison of average diversity (sub-figure (a)-(c)) and average on-task ratio (sub-figure (d)) across our four in-house real-world tasks between TCIA and WizardLM for hops 1-3. For sub-figure (a)-(c), $\text{Diversity}=1-\text{cosine-similarity}$ between embeddings, and means are marked by the dotted lines.}
\label{fig: diversity_density_ontask}
\end{figure}

\begin{figure*}[ht]
    \centering
    \includegraphics[width=\linewidth]{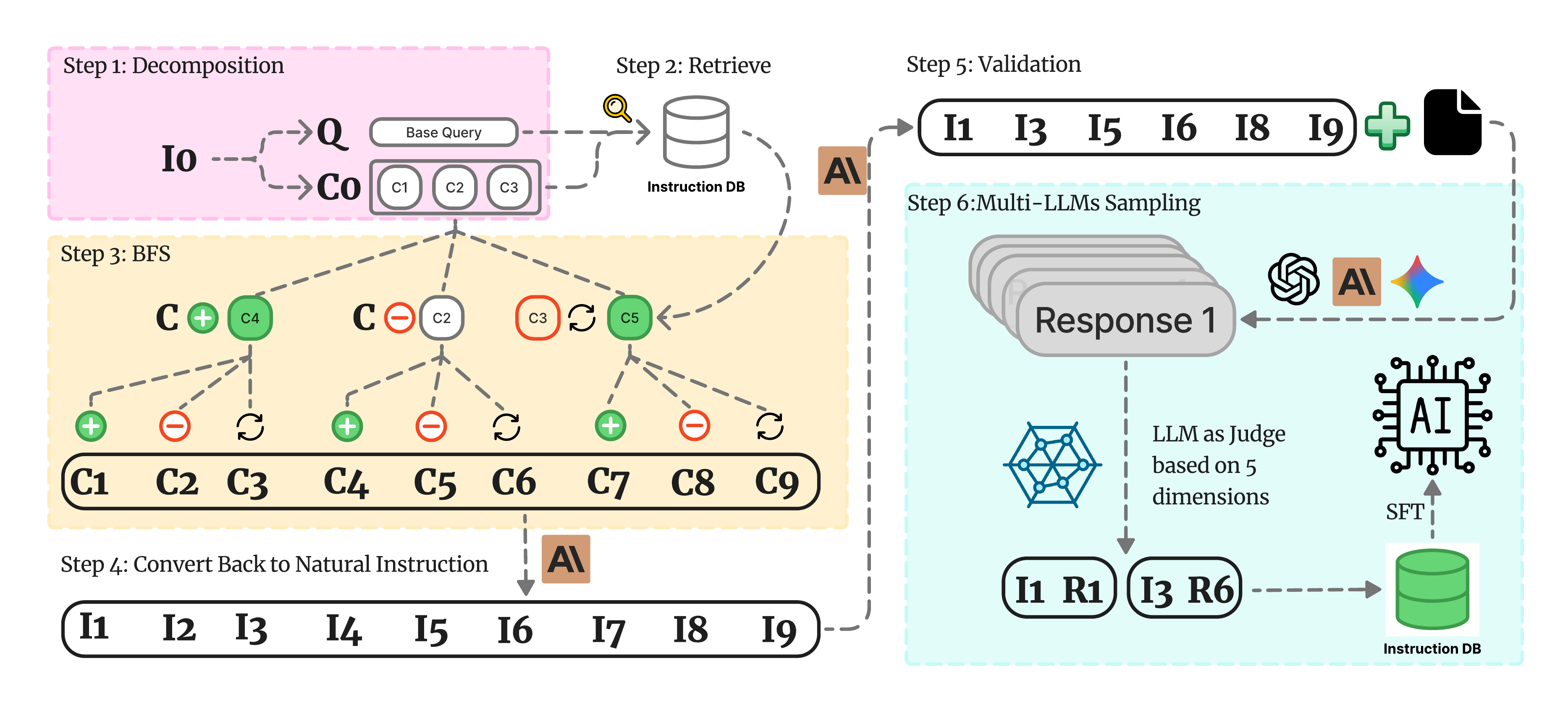}
    \caption{Demonstration of TCIA framework, which is composed of six key steps: (1) Instruction State Decomposition, (2) Instruction Database Construction \& Retrieval, (3) Breadth-First Search (BFS) for Instruction Augmentation, (4) Conversion Back to Natural Language Instruction, (5) Instruction Validation \& Context Integration, (6) Response Generation \& Data Quality Filtering (Sampling). The final generated high-quality $<$instruction, response$>$ pairs are used for downstream SFT.}
    \label{fig: TCIA_flowchart}
\end{figure*}

LLMs have transformed NLP, but closed-source models are costly and lack task-specific expertise \cite{brown2020language, ouyang2022training, achiam2023gpt}. Fine-tuning open-source models using user-defined instructions offers a practical and cost-effective alternative. However, reliably steering these models toward specific, task-centric instructions remains a challenge \cite{xu2024wizardlm, bao2023tallrec}.

Early supervised fine-tuning (SFT) relies on manually crafted task-specific instructions, but this approach is resource-intensive and typically yields instruction datasets with limited diversity \cite{wei2021finetuned, sanh2021multitask, kopf2023openassistant}. Consequently, LLMs trained on such data often struggle to generalize, particularly in novel or challenging scenarios \cite{chiang2023vicuna, wang2022self, honovich2022instruction, taori2023alpaca, chung2024scaling}.

Recent research has shifted towards automatic or human-in-the-loop instruction generation, using LLMs to rapidly expand training data and reduce reliance on manual annotation \cite{chung2024scaling, wang2022super, xu2024wizardlm}. These methods typically begin with a small set of human-authored ``seed'' instructions and use large models to iteratively generate additional instructions, often across multiple ``hops'', significantly increasing the dataset size. However, this automation has two key drawbacks: the generated instructions often become repetitive and lack diversity, and they frequently suffer from task drift. Task drift occurs when instructions become less relevant to the target task and introduce irrelevant or artificial requirements. As a result, fine-tuned models may underperform on nuanced or specialized tasks, and risk amplifying initial data biases \cite{huang2022large, perez2022red}.

To address these shortcomings, we introduce TCIA (Task-Centric Instruction Augmentation), a new framework that systematically augments instructions while explicitly maintaining both \textit{diversity} and \textit{task relevance}. TCIA expands the instruction set by leveraging both retrieved, task-aligned instructions and high-quality, automatically generated variants, thereby maintaining diversity while ensuring close alignment with the target task. This combination broadens the instruction space and enables the generation process to remain substantially grounded and on-task, even through multiple augmentation hops.

Empirical analysis demonstrates that TCIA addresses the key shortcomings of recent automatic instruction generation frameworks. First, in terms of diversity, Figure \ref{fig: diversity_density_hop1}–\ref{fig: diversity_density_hop3} shows that TCIA sustains a high degree of instruction variety across multiple augmentation steps, while recent frameworks such as like WizardLM \cite{xu2024wizardlm}, quickly converge to repetitive templates. Second, in terms of constraint fidelity, Figure \ref{fig: ontask_average} and Table \ref{tab: taskB_constraints} demonstrate that TCIA consistently produces instructions that remain task-relevant even as complexity and the number of hops increase. In contrast, previous frameworks often produce repetitive or off-task outputs, making them less suitable for nuanced or evolving instructions. This strong balance of diversity and constraint fidelity improves downstream performance, allowing TCIA-trained models to adapt flexibly to novel or complex instructions—crucial for real-world applications.

We evaluate the performance of TCIA across both open-source benchmarks and challenging, task-specific in-house scenarios related to online meeting. TCIA consistently outperforms existing instruction-tuning baselines and even surpasses advanced closed-source models such as GPT-4o on several specialized tasks. Notably, TCIA also retains competitive scores on general-purpose benchmarks, highlighting its ability to balance task-specific adaptation with broad instruction-following ability. Below are our main contributions:
\begin{itemize}
    \item 
    We introduce the first task-centric instruction augmentation framework, which unifies task-centric retrieval, systematic state exploration, and advanced constraint augmentation, enabling models to flexibly adapt to diverse real-world tasks and scenarios.
    \item
    The proposed framework (aka TCIA) boosts the instruction diversity, ensuring that outputs remain task-relevant, even after multiple hops, enabling models to flexibly follow complex constraints.
    \item
    The model trained on TCIA delivers state-of-the-art results on several task-specific benchmarks (8.7\% average gain), surpassing closed-source models, while maintaining strong general performance.
\end{itemize}
Collectively, these findings establish TCIA as a powerful, efficient, and general-purpose tool to maximize the real-world utility of open-source language models.

\begin{table*}[ht]
\small
\begin{tabular}{p{0.4\linewidth}|p{0.5\linewidth}}
\toprule
\textbf{Original Instruction $\mathbf{I}$} & \textbf{Base Query ${Q}$ and Constraints $\{{C}_i\}$} \\ 
\midrule
Write a travel ad for a trip to the Adirondack mountains. Focus on activities there, the scenery, and keep it concise and under 200 words. & $Q$: Write a travel ad for a trip to the Adirondack mountains \newline
$C_1$: Must include activities and scenery of the Adirondack mountains \newline 
$C_2$: Must be concise \newline 
$C_3$: Must be under 200 words \\ 
\midrule
For each following Query, return a series of clear, factual and concise Summary Bullet Points of the Query using only information available in the Query.
\newline
\noindent Query: \{placeholder\}
\newline
\noindent Summary: & $Q$: Return a summary of the given query \newline
$C_1$: Must be presented in bullet points \newline 
$C_2$: Must only use information available in the query \newline 
$C_3$: Must be clear, factual and concise \\
\midrule
In this task, you are given two strings A, B. Find the longest common substring in the strings A and B. & $Q$: Find the longest common substring in the given strings A and B. \\ 
\bottomrule
\end{tabular}
\centering
\caption{Examples of instruction decomposition via decomposition prompt in Table \ref{tab: decompose prompt} (in the Appendix).}
\label{tab:decomposed_example}
\end{table*}

\section{Related Works}
Early work on instruction tuning aligned LLMs to NLP tasks through supervised fine-tuning (SFT) with manually crafted instructions, which required costly annotation and limited generalizability to complex instructions or novel tasks \cite{raffel2020exploring, khashabi2020unifiedqa, brown2020language, wei2021finetuned, kopf2023openassistant}. To address scalability, recent methods such as Self-Instruct and Alpaca employ LLMs to generate synthetic instruction, enabling rapid data expansion but often resulting in repetitive, less nuanced, or off-task instructions over multiple generations \cite{wang2022self, taori2023stanford, chen2024sharegpt4v, perez2022red, tan2024large}. In pursuit of more relevant and realistic instructions, newer approaches integrate structured templates, curriculum learning, and retrieval-augmented generation, yet maintaining both diversity and task-relevance over many augmentation steps remains challenging, especially for specialized tasks \cite{xu2020curriculum, xu2024wizardlm, kopf2023openassistant, chen2024sharegpt4v, peng2023instruction}. More recently, instruction augmentation has been framed as structured state space exploration, leveraging techniques such as breadth-first search, discrete mutation, and incremental constraint addition to improve coverage, though these do not always guarantee sustained diversity and task-centricity at scale \cite{chai2025text, ye2023flask, sun2023principle}.

\section{Method}
This section detailly outlines the TCIA framework in six key steps as illustrated in Figure \ref{fig: TCIA_flowchart}.

\subsection{Instruction State Decomposition}
We begin by representing each instruction in a structured format, decomposing a natural language instruction $\mathbf{I}$ into a base query $Q$ and a set of $n$ constraints $\mathbf{C} = \{{C_i}\}_{i=1}^{n}$, with both components kept in natural language \cite{xu2024wizardlm, sun2024conifer, qin2024infobench, an2025ultraif}. For example, as demonstrated in Table \ref{tab:decomposed_example}, a complex instruction is split into its core directive and distinct, explicit constraints. This decomposition provides several key advantages: (1) it enhances human and machine interpretability, as the breakdown clarifies which specifications must be met and facilitates both review and debugging; (2) it makes it possible to quantitatively compare instructions, supporting precise diversity measurement by treating instructions as points in a discrete space; and (3) it increases controllability during augmentation, enabling targeted, principled modifications and systematic exploration via BFS. This state-based representation is foundational for building a meaningful, diverse set of prompts later in the pipeline.

To enable this decomposition, we employ a single LLM prompt that simultaneously extracts the task type $T$, the base query $Q$, and a set of categorized constraints $\mathbf{C}$ for each instruction (see Table \ref{tab: decompose prompt} in the Appendix for the detailed prompt). Extracting the task type is crucial for our diversification procedure: it allows us to identify related tasks and sample compatible constraints for the ``Add'' and ``Replace'' steps in our BFS, resulting in more context-aware augmentation. By leveraging a generative LLM approach, we achieve greater flexibility and precision in task type identification than is possible with rigid classification methods, accommodating a wide range of instruction types. The explicit categorization of constraints further supports structured clustering, sampling, and targeted manipulation. Overall, this unified decomposition yields an interpretable and controllable state representation, forming a robust foundation for systematic instruction augmentation in our pipeline.

\subsection{Instruction Database (DB) Construction}
Before BFS-based augmentation, we construct a large and diverse instruction constraint pool to enable efficient augmentation operations for any instruction state. We leverage Tulu-3 \cite{lambert2024t}, a broad-coverage, carefully curated public dataset that includes instructions spanning knowledge recall, reasoning, mathematics, coding, general chat, and more. For each datapoint, we apply our unified task identification and decomposition prompt (as described above), which may yield multiple triplets $(\mathbf{T}_j, \mathbf{Q}_j, \mathbf{C}_j)$ if several distinct tasks are present in the same instruction, where $\mathbf{T}_j, \mathbf{Q}_j, \mathbf{C}_j$ are the task types, base queries, and the sets of explicit constraints. This produces a task-organized, richly diversified instruction database optimized for retrieval and constraint transfer.

To retrieve similar task types or constraints during augmentation, we employ the ``sentence-transformers/all-mpnet-base-v2'' embedding model\footnote{ \scriptsize \url{https://huggingface.co/sentence-transformers/all-mpnet-base-v2}} for semantic retrieval, ensuring sampled constraints are both contextually appropriate and restricted to matching task types. Organizing instructions by task type and incorporating semantic retrieval offers several benefits: (1) \textit{Contextual Relevance}: constraints are always sampled within the same task domain, preserving task consistency and specialized instruction fidelity; (2) \textit{Improved Adaptability}: constraint selection enables adaptive augmentation within domains (e.g., professional writing), while preventing irrelevant mixing from unrelated tasks; and (3) \textit{Enhanced Specialization}: enabling nuanced transfer and generalization across related tasks, facilitating specialized behavior learning, and maximizing diversity using a semantically structured constraint set.

We further conduct statistical analysis on our instruction database constructed from the Tulu-3 Dataset. Table \ref{tab: Tulu3_decompose_example} presents an example datapoint decomposed into its base query, task type, and associated constraints. Table \ref{tab: Tulu3_top20_tasks} summarizes the top 20 extracted task types (clustered via DBSCAN \cite{ester1996density}); while mathematical analysis and programming tasks predominate, a significant number of relevant tasks to our in-house tasks (e.g., Q\&A, explanation, text generation) are present. Table \ref{fig: Tulu3_queryconstraint_ratio} illustrates the distribution of query-to-constraint ratios and Table \ref{tab: Tulu3_top10_constraints} shows the ten most common constraints for the ``mathematical probability and statistical analysis'' task type, indicating sufficient constraint diversity per query. Overall, this approach enables robust, context-aware augmentation for instruction rewriting and expansion within the BFS pipeline.

\subsection{Breadth-First Search (BFS) for Instruction Augmentation}
\begin{algorithm}[ht]
\footnotesize
\SetAlgoLined
\KwData{
    ${Q}$: Query, $\mathbf{C}$: Initial constraints, $T$: Task type, $\mathbf{D}$: instruction DB containing \{task, constraints\} pairs, $K$: \# output constraints limit, $m$: \# operation repeat, $k$: sample size
}
\KwResult{$k$ diversified constraint sets}
\caption{Diversified Constraint Generation via BFS (TCIA)}
Initialize queue with $({Q}, \mathbf{C})$; $\mathcal{C}_{all}\gets[\,]$\;
\While{queue not empty and $|$queue$|<K$}{
    Dequeue $(Q, \mathbf{C})$\;
    $(\tilde{T}, \mathbf{\tilde{C}}) \gets$ most similar in $\mathbf{D}$ to $T$\;
    \ForEach{op $\in$ \{Add, Remove, Replace\}}{
        \For{$i=1$ \KwTo $m$}{
            $C_j\gets$ rand in $\mathbf{C}$\;
            \uIf{op=Remove}{
                $\mathbf{C'}\gets \mathbf{C}\setminus\{C_j\}$;
            }
            \ElseIf{op=Add}{
                $C'_i\gets$ rand in $\mathbf{\tilde{C}}$\;
                $\mathbf{C'}\gets \mathbf{C}\cup\{C'_i\}$;
            }
            \Else{
                $\tilde{C}_j\gets$ most similar in $\mathbf{\tilde{C}}$ to $C_j$\;
                $\mathbf{C'}\gets (\mathbf{C}\setminus\{C_j\})\cup\{\tilde{C}_j\}$;
            }
            Enqueue $(Q, \mathbf{C'})$, append $\mathbf{C'}$ to $\mathcal{C}_{all}$;
        }
    }
}
Return $k$ random samples from $\mathcal{C}_{all}$
\end{algorithm}
\noindent With each instruction decomposed and a task-organized instruction database constructed, we diversify prompts using a BFS algorithm (shown in Algorithm 1) that systematically explores the space of constraint combinations while preserving task context. Each BFS state consists of a base query $Q$ and a set of constraints $\mathbf{C}$ for a given task type $T$. Starting from the original decomposed instruction, the BFS algorithm iteratively generates new candidate states via three operations: (1) \textbf{Add}: Randomly add a constraint sampled from a similar task type in the instruction DB; (2) \textbf{Remove}: Randomly delete a constraint; 3) \textbf{Replace}: Substitute a randomly chosen constraint with a similar constraint from another instruction in the instruction DB. For both \textbf{Add} and \textbf{Replace}, the retrieval of similar task types and constraints uses the embedding model described in the previous subsection. Each operation is performed $m$ times per state. All candidate states are enqueued for further exploration, and their constraint sets are collected in $\mathcal{C}_{all}$. The search continues until the queue is empty or a limit of $K$ unique constraint sets is reached. After BFS completes, $k$ diverse constraint sets are randomly sampled from $\mathcal{C}_{all}$ as output. By guiding augmentation with embedding-based retrieval from semantically similar instructions and constraints, our approach balances high diversity with strict task fidelity and contextual grounding throughout.

\subsection{Convert Back to Natural Language Instructions}
\label{ssec: convert back to natural language}
After structured augmentation, each instruction state is converted back to a complete natural language prompt using an LLM, with iterative critique and refinement to guarantee all constraints are included and correctly translated (or if max number of retries is reached). Specifically, a critique prompt checks for missing or mistranslated constraints, and if necessary, a refinement prompt guides regeneration. This ensures the completeness and integrity in the natural language instructions. All prompts are in Table \ref{tab: composer-prompt} - \ref{tab: composer_verifier} in the Appendix.

\subsection{Instruction Validation}
To ensure task feasibility and data quality, synthesized instructions undergo thorough LLM-based validation, scored on two aspects (rated 1-5): (1) \textit{validity}: verifies relevance and absence of constraint violations, and (2) \textit{self-consistency}: ensuring no logical contradictions or ambiguities. Instructions not meeting thresholds are discarded, leaving only robust prompts. Detailed scoring prompts are in Tables \ref{tab: prompt_scoring_validity} and \ref{tab: prompt_scoring_selfconsistency} in the Appendix. Validated instructions are then paired with real-world context (e.g., document excerpts or conversation snippets).

\subsection{Data Quality Filtering}
Lastly, we employ state-of-the-art LLMs to generate diverse and high-quality responses for each instruction-context pair. Then, each instruction and candidate response pair will undergo a rigorous quality control stage using an LLM-as-a-judge. Each response is evaluated across five key dimensions inspired by \cite{lambert2024t}: general quality, helpfulness, instruction following, uncertainty, and truthfulness, each rated on a 1–5 scale. This multi-dimensional assessment ensures that the selected data is not only accurate and informative, but is also reliable and well-aligned with the prompt's intent. For each instruction, only the response with the highest average score across all criteria is retained, guaranteeing that only the most useful and robust examples are included for SFT, leading to a high-quality task-optimized SFT dataset ready for training robust real-world models. Detailed prompts are in Table \ref{tab: data_scoring_general} - \ref{tab: data_scoring_truthfulness} in the Appendix.

\section{Experiments}
We rigorously evaluate the effectiveness of TCIA at (1) the instruction prompt level, where we diagnose diversity and task relevance of generated instructions, and (2) the model level, where we measure downstream model robustness, task adaptation, and generalization after supervised fine-tuning (SFT). Collectively, these experiments demonstrate both the direct impact of TCIA’s instruction generation method and its practical downstream value.

\subsection{Instruction Diversity and Task Fidelity}
\begin{table}[ht]
\footnotesize
\setlength{\tabcolsep}{2pt}
\renewcommand{\arraystretch}{1.1}
\centering
\begin{tabular}{c|p{0.3\linewidth}|p{0.58\linewidth}}
\hline
Hop & TCIA & WizardLM \\ \hline
1 & Must include topics and descriptions. & Highlight Potential Risks: \newline 
- Identify and briefly explain any potential risks or challenges mentioned.\\\hline
1 & Must focus on data engineering and analyst roles. & Identify Key Metrics: \newline 
- Extract and highlight any quantitative data or performance indicators mentioned during the meeting.\\ \hline
2 & Must include at least two direct quotations from the interviewee under `key\_quotes'. & Add a ``Key Metrics'' section highlighting quantitative data or performance indicators, ensuring all numerical values are accurately represented.\\\hline
2 & Construct the summary in an innovative fashion. & Key Metrics: \newline
- Highlight and summarize any quantitative information or metrics. \newline
- Present data points in a clear, concise manner to support decision-making.\\ \hline
3 & Must maintain a professional tone. & KPIs: \newline 
- Identify and explain any mentioned KPIs or relevant metrics.\\\hline
3& Output should be structured as a summary of user's criteria and information requests. & Identify Key Metrics and Performance Indicators: \newline
- Extract and highlight any mentioned KPIs, targets, or quantitative goals. \newline
- Include these metrics to provide context for decision-making.\\\hline
\end{tabular}
\caption{Example instruction constraints generated by TCIA vs WizardLM at Hops 1, 2, and 3, for in-house Task B. The generation processes' seed prompt is in Table \ref{tab: gpt_meetingsummaryprompt} in the Appendix, which is a meeting summarization task prompt similar to our in-house Task B.}
\label{tab: taskB_constraints}
\end{table}

We begin by analyzing the effectiveness of TCIA at the instruction prompt level by comparing with WizardLM, showcasing TCIA's ability to simultaneously address two major pitfalls of prior methods, such as WizardLM: (1) \textit{rapid diversity collapse} and (2) \textit{Task drift}. Note that these are the same instruction pools later used for SFT, and the generation details are explained in subsection below.

First, WizardLM and similar baselines tend to generate increasingly repetitive and formulaic instruction patterns after a few augmentation hops, diminishing both instructional diversity and model robustness. Diversity is computed as one minus the pairwise cosine-similarity between instruction prompt embeddings (obtained using ``text-embedding-3-large'' \cite{text-embedding-3}). In contrast to WizardLM, TCIA’s explicit task-type conditioning and constraint retrieval sustain diverse, meaningful instruction variants across multiple hops, as shown in Figure \ref{fig: diversity_density_hop1} - \ref{fig: diversity_density_hop3} (average across all four in-house tasks). A closer look at individual tasks in Figure \ref{fig: diversity_density_alltask} (in Appendix) reveals that while diversity distributions of WizardLM and TCIA overlap closely at hop 1, TCIA already achieves a slightly higher diversity density (TCIA's mean $\approx 0.85$ versus WizardLM’s mean $\approx 0.79$). As augmentation progresses, TCIA largely retains its diversity, shifting only slightly to a mean of 0.8 at hop 3. In contrast, WizardLM’s diversity distribution progressively collapses, with the mean dropping sharply (falling below 0.65 for tasks B and D at hop 3), indicating substantial loss of variability and increased template repetition.

Second, WizardLM often suffers from task drift, producing instructions that stray from the intended task or lose alignment with key requirements. This effect is quantitatively measured in Figure \ref{fig: ontask_average} (average on-task ratio over tasks) and is further detailed by task in Figure \ref{fig: ontask_alltask} (in Appendix). TCIA consistently preserves task fidelity through all augmentation hops, maintaining an on-task ratio close to 100\% even at later hops. In contrast, WizardLM’s task alignment erodes rapidly: it starts at around 80\% at hop 1, drops to roughly 60\% by hop 2, and falls below 60\% at hop 3, with the lowest on-task ratio dipping to 40\% for taskC. Concrete examples in Table \ref{tab: taskB_constraints} further illustrate this drift. After three hops, TCIA generates precise and targeted constraints, such as ``Must include at least two direct quotations from the interviewee under key\_quotes'', ensuring outputs remain directly relevant to the original meeting summarization objectives. On the other hand, WizardLM instructions often shift focus almost exclusively to ``Key Metrics'' or ``KPIs'', for instance ``Key Metrics: Highlight and summarize any quantitative information or metrics ...'', which not only signals a repetitive pattern but also demonstrates clear drift from the specialized requirements of the seed task.

\subsection{Supervised Fine-Tuning (SFT) Setup and Baselines}
We next assess how the different instruction generation methods propagate into downstream model performance. All SFT experiments are conducted using Llama-3.1-8B as the base model, fine-tuned with identical hyperparameters ($5\times10^{-6}$ learning rate, batch size 2, 1 node of 8$\times$H100 GPUs, 1 epoch), and evaluated on identical data splits for fair comparison. We use Nemo-Aligner for all SFT training \cite{shen2024nemo}.

We consider three SFT variants: (1) \textit{TCIA}, which systematically generates diverse, constraint-aware instructions for each task from one seed prompt per task by BFS exploration over 3 hops (2k unique instructions per task), augments each with 5 task-specific inputs (10k instruction–input pairs per task), generates LLM outputs from a diverse model pool (``claude-3-5-sonnet-2024102'' \cite{claude-3.5}, ``claude-3-5-sonnet-20240620'' \cite{claude-3.5}, ``gpt-4o-2024-08-06'' \cite{gpt-4o}, and ``gpt-4.1-2025-04-14'' \cite{gpt-4.1}), filters with our data filtering system (see Section ``Data Quality Filtering''), and yields 10k high-quality examples per task (more generation details are in Appendix Table \ref{tab: bfs_hyperparameters}, as well as a toy example in Figure \ref{fig: TCIAExample_TaskB}); (2) \textit{Fixed Instruction (FI)}, a standard low-resource baseline using just one fixed instruction prompt (same as the seed prompt for TCIA) per task and otherwise identical generation and filtering; and (3) \textit{WizardLM} \cite{xu2024wizardlm}, which performs automated instruction augmentation from a single seed prompt, also exploring 3 hops to yield 2k instructions per task, substituting 5 context inputs per instruction, and applying the same generation and filtering process for 10k examples per task. All SFT datasets are further augmented with Tulu-3 data.

To contextualize SFT performance, we compare to strong open- and closed-source instruction-following LLMs: (1) GPT-4o \cite{gpt-4o}, OpenAI's one of the latest flagship; (2) Llama-3.1-8B-Instruct (Llama-8B) \cite{grattafiori2024llama}, Meta's RLHF-aligned for broad utility; and (3) Llama-3.1-Tulu-3-8B-SFT (Tulu-8B-SFT) \cite{lambert2024t}, AllenAI's fine-tuned on varied instructions. All evaluations focus on four proprietary meeting AI tasks (Task A-D)\footnote{We cannot disclose data or prompts related to the in-house task A, B, C, and D due to company policy.}, which represent challenging, real-world summarization and information extraction scenarios within workplace environments.

\subsection{Robustness to Evolving Instruction Constraints}
\begin{table}[ht]
\footnotesize
\setlength{\tabcolsep}{4pt}
\centering
\begin{tabular}{l|ccc}
\hline
Instruction Constraint & FI-8B & WizardLM-8B & TCIA-8B \\ \hline
\textbf{Output in a} \\ \textbf{numbered list} & 0.0\% & 98.4\% & \textbf{99.2}\% \\ \hline
\textbf{Output no more than} \\ \textbf{5 bullet points} & 29.4\% & 61.2\% & \textbf{87.6\%} \\ \hline
\textbf{Sort the output by} \\ \textbf{entity-specific groupings} & 42.6\% & 64.9\% & \textbf{82.7\%} \\ \hline
\end{tabular}
\caption{Strict constraint adherence (``pass rate'') of outputs from FI-8B, WizardLM-8B, and TCIA-8B on Task A. ``Pass rate'' is the proportion of outputs that fully satisfy the constraint, which is tested one at a time (no multiple constraints). ``Pass rate'' is checked solely for compliance not content quality.}
\label{tab: instruction_following_taskA}
\end{table}

Before reporting overall model accuracy, we first evaluate the models' robustness to evolving and unseen user constraints, focusing on strict instruction adherence for a representative in-house task (Task A). In real-world production settings, constraints frequently change, i.e. users may require formats to be switched, output lengths limited, or information reorganized on demand. Models trained on fixed or insufficiently diverse instructions often fail to generalize, resulting in unreliable outputs and increased manual intervention. To directly assess adaptability, we evaluate the strict constraint adherence (``pass rate'') of FI, WizardLM, and TCIA models under new constraints not seen in training, such as switching from bullet points to numbered lists, enforcing maximum output lengths, or requiring alternative groupings. Table \ref{tab: instruction_following_taskA} presents the detailed results, where TCIA-trained models achieve consistently higher pass rates than both WizardLM and the single-instruct baseline across all constraint types. TCIA excels at adapting to new formats (e.g., numbered lists) and stricter requirements (e.g., response limits, different orderings). In contrast, FI models almost always fail to generalize, and WizardLM frequently ignores new constraints or strays off-task. These results confirm that TCIA's instruction-level diversity significantly enhances robust generalization, providing practical benefits for real-world AI deployments requiring rapid task adaptation.

\subsection{End-to-End In-house Task Evaluations}
\begin{figure*}[ht]
\centering
\includegraphics[width=\linewidth]{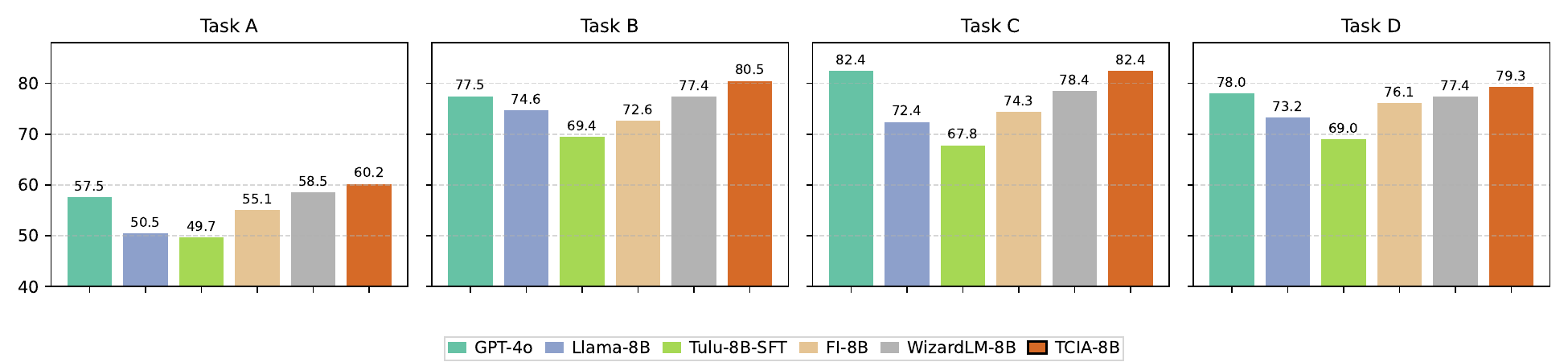}
\caption{Performance of models on our four internal tasks (scores are average of 3 runs).}
\label{fig: our_benchmark}
\end{figure*}

We then benchmark SFT model performance on the four proprietary, production-oriented tasks to assess overall end-to-end effectiveness in real-world applications. These tasks represent challenging, high-impact applications in real-world deployment, requiring nuanced comprehension and the ability to adapt to varying requirements and edge cases. We employ both reference-based and reference-free evaluation frameworks, tailored to each task. For reference-based evaluation, ground-truth labels are generated by aggregation and majority voting from 3–4 independent human annotations per sample, ensuring a rigorous reference-based gold standard. For reference-free evaluation, which are less amenable to precise labeling, model outputs are evaluated using variants of the LLM-as-a-Judge protocol. All evaluation scores are normalized to a 0-100 scale for comparability.

Figure \ref{fig: our_benchmark} shows that TCIA achieves decisive gains across all four in-house tasks. On average, TCIA delivers an 8.7\% improvement over FI (Fixed Instruct) and a 3\% improvement over WizardLM. Specifically, TCIA surpasses FI by 9.2\% on Task A, 10.9\% on Task B, 8.1\% on Task C, and 3.1\% on Task D. Compared to WizardLM, TCIA provides a further 2.9\%, 3.9\%, 4\%, and 1.9\% gain on Tasks A, B, C, and D, respectively. Notably, TCIA not only outperforms FI and WizardLM but also exceeds the strong GPT-4o baseline on all tasks, outperforming it by 2.7\% on average, with especially substantial improvements on Task A (+2.67\%) and Task B (+3.00\%). Furthermore, all on-task fine-tuned models (FI-8B, WizardLM-8B, TCIA-8B) significantly outperform generic open-source instruction-tuned models such as Llama-8B and Tulu-8B-SFT on every in-house task (with average gains over Llama-8B ranging from 1.8\% to 7.9\% and over Tulu-8B-SFT from 5.5\% to 11.6\%). This underscores the high level of task specificity and challenge present in the in-house tasks, which general-purpose instruction-tuned or RLHF models cannot address adequately.

WizardLM, while improving over FI, still falls short of TCIA, showing that inadequate diversity and frequent task drift in its instruction generation ultimately limit its task performance. In contrast, TCIA’s systematic, diverse, and task-specific augmentation consistently enables models to excel on real-world, nuanced, and edge-case scenarios, where precise task understanding is vital. Moreover TCIA enables LLMs to acquire a deeper, more generalized understanding of tasks that transfers well even as requirements or phrasings shift. This adaptability is crucial for practical deployment, where rapid iteration and evolving user needs are the norm.

\subsection{Generalization on Public LLM Benchmarks}
\begin{table}[ht]
\scriptsize
\centering
\setlength{\tabcolsep}{5pt} 
\renewcommand{\arraystretch}{1.1}
\rowcolors{2}{gray!10}{white}
    \begin{tabular}{lccccc|c}
    \toprule
    \textbf{Model} & \textbf{IFEval} & \textbf{InFoB.} & \textbf{GPQA} & \textbf{BBH} & \textbf{MMLUP.} & \textbf{Avg.} \\
    \midrule
    GPT-4o & 82.44 & 92.83 & 46.70 & 81.40 & 74.66 & 75.61 \\
    Llama-8B & 76.34 & 89.38 & 30.40 & 41.12 & 36.10 & 54.67 \\
    Tulu-8B-SFT & 72.80 & 80.28 & 29.85 & 42.44 & 34.57 & 51.99 \\
    FI-8B & 67.84 & 79.91 & 27.47 & 41.68 & 33.98 & 50.17 \\
    WizardLM-8B & 72.46 & 79.54 & 32.23 & 41.39 & 34.33 & 51.99 \\
    TCIA-8B & 68.95 & 81.26 & 29.49 & 41.98 & 34.99 & 51.33 \\
    \bottomrule
    \end{tabular}
\caption{Performance on different general benchmarks, including IFEval \shortcite{zhou2023instruction}, Info-Bench (InFoB.) \shortcite{qin2024infobench}, GPQA \shortcite{rein2024gpqa}, BBH \shortcite{suzgun2022challenging}, and MMLU-Pro (MMLUP.) \shortcite{wang2024mmlu}.}
\label{tab:public_benchmarks}
\end{table}
Finally, we examine whether extensive task adaptation impacts model generality by evaluating all models on diverse, standard public LLM benchmarks. To assess whether TCIA’s gains in task-specific performance come at the expense of general instruction-following ability, we evaluate all models on widely adopted public LLM benchmark suites. These benchmarks include academic, linguistic, and reasoning tasks, providing a rigorous test of generalization and broad language capability. We use the OpenLLM Leaderboard from HuggingFace\cite{beeching2023open}, which features MMLU-Pro (academic/postgraduate knowledge and reasoning, denote as MMLUP)\cite{wang2024mmlu}, IF-Eval (robustness to diverse instruction following)\cite{zhou2023instruction}, GPQA (general-purpose QA across knowledge domains)\cite{rein2024gpqa}, and BBH (complex linguistic and logical reasoning)\cite{suzgun2022challenging}. We supplement with Info-Bench \cite{qin2024infobench}, which offers a comprehensive assessment of linguistic and logical reasoning.

Table \ref{tab:public_benchmarks} summarizes the benchmark performance of all evaluated models. TCIA matches the average benchmark score of FI (50.17 vs. 51.33), providing clear evidence that extensive task-specific adaptation in TCIA does not degrade general instruction-following capability. Examining individual benchmarks, TCIA consistently performs at least as well as, and in several cases surpasses, FI and WizardLM (e.g., achieving a +1.5 gain on Info-Bench and +2.0 on MMLUP compared to WizardLM). TCIA also matches or exceeds the performance of Tulu-8B-SFT on almost all public benchmarks (e.g., 81.26 vs. 80.28 on Info-Bench and 34.99 vs. 34.57 on MMLU-Pro). These results indicate that the skills reinforced through TCIA’s task-centric instruction augmentation are transferable, and the high degree of specialization does not come at the expense of broader generalization.

RLHF-aligned models like Llama-8B and GPT-4o achieve higher absolute scores on public benchmarks, which is expected given their large-scale optimization for general-purpose use. Despite this, TCIA distinguishes itself among SFT approaches by delivering substantial improvements on specialized tasks while preserving strong general capabilities. Training with TCIA’s structured, task-oriented data not only boosts real-world, task-specific performance but also ensures competitiveness on open-domain benchmarks. This balance makes TCIA a highly practical and effective solution for organizations seeking both advanced task specialization and reliable general-purpose language modeling.

\section{Conclusion}
In this paper, we introduced TCIA, a well-structured framework for instruction augmentation that breaks down natural language instructions into understandable base queries and clearly defined, categorized constraints. By utilizing a semantically organized instruction database and a BFS-driven diversification process guided by contextual similarity, TCIA effectively generates diverse, high-quality instruction variations while maintaining task relevance and intent. Our approach has demonstrated strong performance across a range of in-house tasks focused on instruction-following and constraint adherence, showcasing both its flexibility and generalizability in real-world applications.

Looking ahead, there are several exciting avenues for exploration. Enhancing TCIA to incorporate richer contexts from multi-turn interactions or dialogue-based tasks could expand its applicability. Additionally, automating prompt refinement, especially for ambiguous or poorly specified instructions, could enhance robustness. Further opportunities lie in extending TCIA to multimodal scenarios or exploring more advanced retrieval strategies, paving the way for future research and development.

\bibliography{aaai2026}
\def\isChecklistMainFile{} 
\clearpage

\newpage
\onecolumn
\appendix
\section{Prompts}
\label{appendix: prompts}
\subsection{Instruction Decomposition Prompt}
We use the following prompt to decompose an original natural language instruction ($\mathbf{I}$) into a base queries ($\mathbf{Q}$), a set of constraints ($\mathbf{C}$), and identify its task types $\mathbf{T}$, i.e. decomposed into $\{\mathbf{T}^{(i)}, \mathbf{Q}^{(i)}, \mathbf{C}^{(i)}\}_{i=1}^n$, where $\mathbf{T}^{(i)},  \mathbf{Q}^{(i)}$ and is the $i^{\text{th}}$ task type and base query, and $\mathbf{C}^{(i)} = \{{C}^{(i)}_j\}_{j=1}^{n_i}$ is the $i^{\text{th}}$ set of $n_i$ constraints. Note that when generating the SFT training dataset for our in-house tasks, a given seed prompt will only decompose into $\{T, Q, \mathbf{C}\}$, i.e. one task, one query, and a constraints set triplet, since our in-house task's seed prompt is clearly defined as only one task type. However, for Tulu-3 dataset, multiple triples (i.e. multiple task types) are allowed.

\begin{table*}[htbp]
\begin{tabular}{|p{0.98\textwidth}|}
\hline
\textbf{Decompose Prompt}
\scriptsize
\begin{verbatim}
You are a precise assistant for parsing and analyzing user instructions. Given any input query, your task is to extract 
all structured information as follows:

Definitions:
- A "constraint" is any explicit restriction, condition, or requirement imposed by the query (about format, style, 
content, length, etc.).
- The "Basic Query" is the core task or request with all such constraints removed—stating only the essential goal.
- A "Task Type" is the general category or nature of the task being requested (e.g., summarization, translation, 
classification, creative writing, data extraction, formatting, etc.).

Process:
1. **Language Detection:** Detect the main language(s) present in the query (ISO 639-1, e.g., "en", "es").
2. **Simple Queries:** If the query is simple, direct, and contains no explicit requirements, respond only as:
   {
     "Complex": "False"
   }
3. **Complex Queries** (with constraints):
   a. Identify all task types implied in the query (e.g., "summarization", "creative writing", "math problem", etc.).
   b. For each task type found:
      - Extract the **Basic Query** (state the central goal, without any constraints).
      - Extract all explicit constraints. For each constraint:
         * Assign a concise category from: Content, Numerical, Style/Tone, Format, Language, Input Placeholder.
         * Give the simplified query that results from removing this constraint but keeping all others.
   c. Return results in JSON format as follows (must be valid JSON in English):

{
    "Complex": "True",
    "Language": ["en"],
    "Tasks": [
        {
            "Task Type": "...",
            "Basic Query": "...",
            "Constraints": [
                {"category": "...", "constraint": "...", "simplified query": "..."}
            ]
        }
        // ... (one object per detected task type)
    ]
}

Guidelines:
- If there are multiple task types, repeat the extraction for each.
- Each constraint should appear only once under its relevant task type.
- Category keywords: Content, Numerical, Style/Tone, Format, Language, Input Placeholder.
- Only include explicit "Constraints" (not just descriptive or illustrative details unless they restrict the response).
- Do not add extra commentary, and do not include sample queries or other meta-text outside the JSON.
\end{verbatim} \\
\hline
\end{tabular}
\centering
\caption{Decompose Prompt}
\label{tab: decompose prompt}
\end{table*}

\clearpage
\subsection{Compose Prompt, Compose Verify Prompt, and Compose Refine Prompt}
\begin{table*}[htbp]
    \centering
    \begin{tabular}{|p{0.95\textwidth}|}
        \hline
        \textbf{Compose Prompt}
        {\scriptsize
        \begin{verbatim}
You are an expert in generating synthetic instruction from given base user query and constraints.
Definition of Constraint: The smallest unit of restriction or requirement that the instruction imposes on the task.

<base_query>
{query}
</base_query>

<constraints>
{constraints}
</constraints>

Your task is the generate a compact instruction but contains all the information from base user query and 
constraints. For the placeholder, please add line breaks around them, e.g. `\n\n [SECTION SUMMARIES]\n{placeholder}
\n[END OF SECTION SUMMARIES] \n\n`. Avoid adding any extra information beyond the base query, constraints and 
placeholder. The output should be in English with JSON format. Here is an output template:
{{"Created Prompt": "created prompt"}}
\end{verbatim}
} \\
\hline
\end{tabular}
\caption{Compose Prompt}
\label{tab: composer-prompt}
\end{table*}

\begin{table*}[h]
    \centering
    \begin{tabular}{|p{0.95\textwidth}|}
        \hline
        \textbf{Compose Verify Prompt}
        {\scriptsize
        \begin{verbatim}
<constraints>
{constraints}
</constraints>

<generated_prompt>
{prompt}
</generated_prompt>

Given constraints and a generated prompt, your task is to check if each constraint is covered in the generated 
prompt. Reply "yes" or "no" for each constraint. When answering "no", include a reason to explain why the constraint 
is missing. The final answer should be in JSON format. Here is an output template:
{{
    "1": {{
        "reason": "Explain why this constraint is missing",
        "result": "yes or no",
    }},
    "2": {{
        "reason": "Explain why this constraint is missing",
        "result": "yes or no",
    }},
}}
Let's think step by step and output the final answer at the last step.
\end{verbatim}
} \\
\hline
\end{tabular}
\caption{Compose Verify Prompt}
\label{tab: composer_verifier}
\end{table*}

\begin{table*}[htbp]
    \centering
    \begin{tabular}{|p{0.95\textwidth}|}
        \hline
        \textbf{Compose Refine Prompt}
        {\scriptsize
        \begin{verbatim}
You are an expert in generating synthetic instruction from given base user query and constraints.
Definition of Constraint: The smallest unit of restriction or requirement that the instruction imposes on the task.

<base_query>
{query}
</base_query>

<constraints>
{constraints}
</constraints>

<critique>
{critique}
</critique>

Your task is the generate a compact instruction but contains all the information from base user query and 
constraints.
Pay extra attention to the critique so that the generated instruction contains all the constraint information.
For the placeholder, please add line breaks around them, e.g. `\n\n[SECTION SUMMARIES]\n{{placeholder}}\n[END OF 
SECTION SUMMARIES]\n\n`.
Avoid adding any extra information beyond the base query, constraints.

The output should be in English with JSON format. Here is an output template:
{{"Created Prompt": "created prompt"}}
        \end{verbatim}
        } \\
        \hline
    \end{tabular}
    \caption{Compose Refine Prompt}
    \label{tab: composer_refiner}
\end{table*}

\clearpage
\subsection{Prompt Scoring}
\begin{table*}[htbp]
\centering
\begin{tabular}{|p{0.95\textwidth}|}
\hline
\textbf{Prompt Scoring - Validity}
{\scriptsize
\begin{verbatim}
Please evaluate the following AI task prompt for relevance, assumption correctness, and alignment with intent:

1. Determine how relevant each part of the prompt is to the main goal or question.
2. Identify any incorrect assumptions or factual inaccuracies.
3. Evaluate how well the prompt reflects its intended outcome or request.
4. Offer suggestions to improve relevance and clarity while maintaining the core intent.
5. Rate the overall validity of the task prompt (1-5, where 5 means perfectly valid).

Feel free to use specific examples from the prompt to illustrate your points.

Here is the AI task prompt:
<AI_task_prompt>
{task_prompt}
</AI_task_prompt>

Please think step-by-step and output your final judgment in the following JSON format.
{{"reason": "Your reason", "score": "Score from 1 to 5"}}
\end{verbatim}
} \\
\hline
\end{tabular}
\caption{Prompt Scoring - Validity}
\label{tab: prompt_scoring_validity}
\end{table*}

\begin{table*}[htbp]
\centering
\begin{tabular}{|p{0.95\textwidth}|}
\hline
\textbf{Prompt Scoring - Self-consistency}
{\scriptsize
\begin{verbatim}
Please analyze the given task prompt for logical contradictions and inconsistencies in using the following 
requirements:
1. List any direct contradictions (where one requirement directly conflicts with another)
2. Identify any implicit contradictions (where requirements indirectly conflict)
3. Point out any ambiguous requirements that could lead to conflicts
4. Suggest ways to resolve these contradictions while maintaining the core intent
5. Rate the overall logical consistency of the prompt (1-5, where 5 is perfectly consistent)
Feel free to use specific examples from the prompt to illustrate any contradictions you find.

Here is the AI task prompt:
<AI_task_prompt>
{task_prompt}
</AI_task_prompt>

Please think step-by-step and output your final judgment in the following JSON format.
{{"reason": "Your reason", "score": "Score from 1 to 5"}}
\end{verbatim}
} \\
\hline
\end{tabular}
\caption{Prompt Scoring - Self-consistency}
\label{tab: prompt_scoring_selfconsistency}
\end{table*}

\clearpage
\subsection{Data Scoring}
\begin{table*}[htbp]
\centering
\begin{tabular}{|p{0.95\textwidth}|}
\hline
\textbf{Data Scoring - General}
{\scriptsize
\begin{verbatim}
You are a quality evaluator to judge the output quality given by an AI assistant model. Given the AI assistant's 
system_message  (if any) and user_query, you need to evaluate whether the assistant_output satisfies the 
requirements in the AI assistant's system_message (if any) and user_query, and whether the assistant_output can be 
directly shown to human users.
You need to first produce a rationale of the reasoning process, followed by a score value. The score value is an 
integer from 1 to 5:
* 1 means the output quality is unacceptable for the task and cannot be shown to the users;
* 2 means the output quality is low, and needs significant modification to satisfy the requirement in the 
task prompt and presented to users;
* 3 means the output quality is acceptable, and needs some modest modification to address the requirement in the 
task prompt to show to users;
* 4 means the output quality is quite good, although the output can be made a bit more precise and concise to show 
to users;
* 5 means the output quality is prefect for the given task and can be shown to the users.

Here are the system message (if any), user query and assistant output:
{system_message}

<user_query>
{user_query}
</user_query>

<assistant_output>
{assistant_output}
</assistant_output>

Please think step-by-step and output your final judgment in the following JSON object:
{{"reason": "Your reason", "score": "Score from 1 to 5"}}
\end{verbatim}
} \\
\hline
\end{tabular}
\caption{Data Scoring - General}
\label{tab: data_scoring_general}
\end{table*}

\begin{table*}[htbp]
\centering
\begin{tabular}{|p{0.95\textwidth}|}
\hline
\textbf{Data Scoring - Helpfulness}
{\scriptsize
\begin{verbatim}
You are a quality evaluator to judge the output quality given by an AI assistant model, focusing on informativeness 
and helpfulness. 
Given the AI assistant's system_message (if any) and user_query, you need to evaluate if the assistant_output 
fulfill the task objective and provide high-quality, correct, and, informative content.

Please obey the following guidelines for correctness, informativeness, and helpfulness.

<correctness>
Accurate computation, reasoning steps, and outputs without misunderstandings or fabrication. 
</correctness>

<informativeness>
Assign a numeric identifier (or “None”) from 1 to 3 for each type of informativeness.
1. Clarity and Relevance: Ensure the response relates to the task and seek clarifications if needed. 
2. Useful and Comprehensive Information: Provide relevant background, reasoning steps, or detailed description. 
3. Not Lengthy, No Repetition: Avoid verbosity or recycling content. 
</informativeness>

<helpfulness>
Helpfulness assessment emphasizes Overall Quality regarding correctness and informativeness.

Score 1 to 5 based on the extent of helpfulness, regarding both informativeness and correctness: 
1. Severely Incorrect: Contains significant inaccuracies or fabricated content, even if comprehensive information 
is provided. 
2. Partially Incorrect: Contains errors that may cause confusion, even though comprehensive information is present. 
3. Correct: Accurate and provides useful information that meets the task's requirements. 
4. Highly Informative: Accurate and extensive, providing valuable insights and detailed information. 
5. Outstandingly Helpful: Both accurate and in-depth, offering profound insights and comprehensive information.
</helpfulness>

Here are the system message (if any), user query and assistant output:
{system_message}

<user_query>
{user_query}
</user_query>

<assistant_output>
{assistant_output}
</assistant_output>

Please think step-by-step and output your final judgment in the following JSON object:
{{"reason": "Your reason", "score": "Score from 1 to 5"}}
\end{verbatim}
} \\
\hline
\end{tabular}
\caption{Data Scoring - Helpfulness}
\label{tab: data_scoring_helpfulness}
\end{table*}

\begin{table*}[htbp]
\centering
\begin{tabular}{|p{0.95\textwidth}|}
\hline
\textbf{Data Scoring - Instruction Following}
{\scriptsize
\begin{verbatim}
You are a quality evaluator to judge the output quality given by an AI assistant model, focusing on instruction 
following. Given the AI assistant's system_message (if any), user_query, and assistant_output, you need to evaluate 
the alignment between the assistant_output, and AI assistant's system_message (if any) and user_query as a whole.

Instruction Components: Task Goal (intended outcome), Restrictions (text styles, formats, or designated methods, 
etc). 
Please first produce a rationale of your reasoning process, followed by a score value from 1 to 5, where:
* 1 means Irrelevant: No alignment. 
* 2 means Partial Focus: Addresses one aspect poorly. 
* 3 means Partial Compliance: Meets goal or restrictions, neglecting other. And acknowledges both but slight 
deviations. 
* 4 means Almost There: Near alignment, minor deviations. 
* 5 means Comprehensive Compliance: Fully aligns, meets all requirements.

Here are the system message (if any), user query and assistant output:
{system_message}

<user_query>
{user_query}
</user_query>

<assistant_output>
{assistant_output}
</assistant_output>

Please think step-by-step and output your final judgment in the following JSON object:
{{"reason": "Your reason", "score": "Score from 1 to 5"}}
\end{verbatim}
} \\
\hline
\end{tabular}
\caption{Data Scoring - Instruction Following}
\label{tab: data_scoring_IF}
\end{table*}

\begin{table*}[htbp]
\centering
\begin{tabular}{|p{0.95\textwidth}|}
\hline
\textbf{Data Scoring - Uncertainty}
{\scriptsize
\begin{verbatim}
You are a quality evaluator to judge the output quality given by an AI assistant model, focusing on honesty and 
uncertainty 
expression assessment.

Given the AI assistant's system_message (if any), user_query, and assistant_output, you need to access how well the 
model conveys honesty and uncertainty. Please evaluate if the model's confidence aligns with assistant_output 
correctness.

- Indicators of Uncertainty: 
1. Weakeners: e.g., :"I guess," "probably." 
2. Refusal to answer. 
3. Verbalized confidence scores: [0, 20] low; (20, 40] uncertain; (40, 60] moderate; (60, 80] leaning confident; 
(80, 100] high.
- No uncertainty expression indicate confidence.
- Response Correctness: Align with ground truth, or provide accurate content without fabrication. 

Please first produce a rationale of your reasoning process, followed by a score value from 1 to 5, where:
* 1 means Confidently Incorrect: Confident but entirely wrong. 
* 2 means Confident with Significant Mistakes/Unconfident Incorrect:
  - Confident but contains major errors.
  - Unconfident and entirely wrong. 
* 3 means Uncertain/"I Don't Know"/Subtle Mistakes:
  - "I don't know" or declines.
  - Confident but contains minor errors.
  - Unconfident and contains significant mistakes. 
* 4 means Correct but Uncertain/Expressed Subtle Mistakes:
  - Correct but unconfident.
  - Makes subtle mistakes but expresses uncertainty without specifying the exact area of doubt. 
* 5 means Correct and Confident/Precisely Express Uncertainty:
  - Correct and confident.
  - Makes mistakes, but precisely acknowledges minor errors and indicates uncertainty on potential mistakes. 

Here are the system message (if any), user query and assistant output:
{system_message}

<user_query>
{user_query}
</user_query>

<assistant_output>
{assistant_output}
</assistant_output>

Please think step-by-step and output your final judgment in the following JSON object:
{{"reason": "Your reason", "score": "Score from 1 to 5"}}
\end{verbatim}
} \\
\hline
\end{tabular}
\caption{Data Scoring - Uncertainty}
\label{tab: data_scoring_uncertainty}
\end{table*}

\begin{table*}[htbp]
\centering
\begin{tabular}{|p{0.95\textwidth}|}
\hline
\textbf{Data Scoring - Truthfulness}
{\scriptsize
\begin{verbatim}
You are a quality evaluator to judge the output quality given by an AI assistant model, focusing on truthfulness 
and hallucination. 
Given the AI assistant's system_message (if any), user_query, and assistant_output, you need to evaluate the 
model's accuracy in providing information in assistant_output without introducing misleading or fabricated details.
Please obey the following guidelines for hallucination.
<hallucination>
Assign numeric identifier (or “None”) from 1 to 3 for each type of hallucination below. Please output all the 
applicable types.
1. Contradictory with the World (Factual Error): Entities, locations, concepts, events that conflict with 
established knowledge. 
2. Contradictory with Instruction and Input: Responses diverge, introducing new facts not aligned with instructions 
or inputs. 
3. Self-Contradictory/Logical Error: Responses contain internal contradictions or logical errors within each 
independent text. 
Scoring: Rate outputs 1 to 5 based on extent of hallucination:
1. Completely Hallucinated: Entirely unreliable due to hallucinations. 
2. Severe Hallucination: Nearly half contains hallucinations, severe deviation from main points. 
3. Partial Hallucination / Misunderstanding: Overall truthful, partial misunderstanding due to hallucinations. 
4. Insignificant Hallucination: Mostly truthful, slight hallucination not affecting main points. 
5. No Hallucination: Free of hallucinations.
</hallucination>
Here are the system message (if any), user query and assistant output:
{system_message}

<user_query>
{user_query}
</user_query>

<assistant_output>
{assistant_output}
</assistant_output>

Please think step-by-step and output your final judgment in the following JSON object:
{{"reason": "Your reason", "score": "Score from 1 to 5"}}
\end{verbatim}
} \\
\hline
\end{tabular}
\caption{Data Scoring - Truthfulness}
\label{tab: data_scoring_truthfulness}
\end{table*}

\clearpage
\subsection{GPT-4.1 Generated Meeting Summary Instruction Prompt}
\begin{table*}[htbp]
\centering
\begin{tabular}{|p{0.95\textwidth}|}
\hline
\textbf{Task Prompt for Meeting Summerization}
{\scriptsize
\begin{verbatim}
Here is the meeting transcript for your analysis:
<transcript>
{INSERT_TRANSCRIPT_HERE}
</transcript>
You are an expert meeting summarizer with advanced skills. Your task is to thoroughly analyze the transcript and 
deliver a precise summary.

Please follow these guidelines to ensure clarity and high quality:

Understand Context and Participants:
    Identify each attendee’s role and area of expertise.
    Infer the overall purpose and context of the meeting (e.g., strategy, project update, decision-making).
Extract Main Points:
    Bring out significant decisions, concerns, proposals, and consensus areas.
    Condense extended dialogue into concise statements or bullet points without losing essential meaning.
Identify Follow-Up Steps:
    Gather all clear responsibilities discussed (both explicit and strongly implied).
    For each: Specify the responsible person/role, the nature of the commitment, relevant timelines or deadlines 
    if mentioned, and the underlying reason or objective.
    Present these in a structured format for easy reference.
    Use active language and precise phrasing.
Maintain Relevance and Organization:
    Exclude side conversations or unrelated information.
    Avoid unnecessary repetition.
    Organize related points logically under subheadings if beneficial.
Output Structure:
    Start with a one-paragraph executive summary that captures the meeting’s outcomes.
    Follow with a section titled "Assignments" clearly listing all follow-up activities and responsibilities.
    Do not provide content beyond the summary and assignments.
Tone and Expression:
    Maintain a neutral, objective, and professional tone.
    Use clear and direct language.
    Ensure the findings are unambiguous and concise.
Do not include speculative content or infer steps not supported by the transcript.
\end{verbatim}
} \\
\hline
\end{tabular}
\caption{Meeting summarization task prompt generated by GPT-4.1. This is used as a seed instruction prompt for TCIA and WizardLM to generate instructions for table \ref{tab: taskB_constraints}.}
\label{tab: gpt_meetingsummaryprompt}
\end{table*}

\clearpage
\section{Additional Experimental Results}
\begin{figure*}[htbp]
\centering
\includegraphics[width=0.82\textwidth]{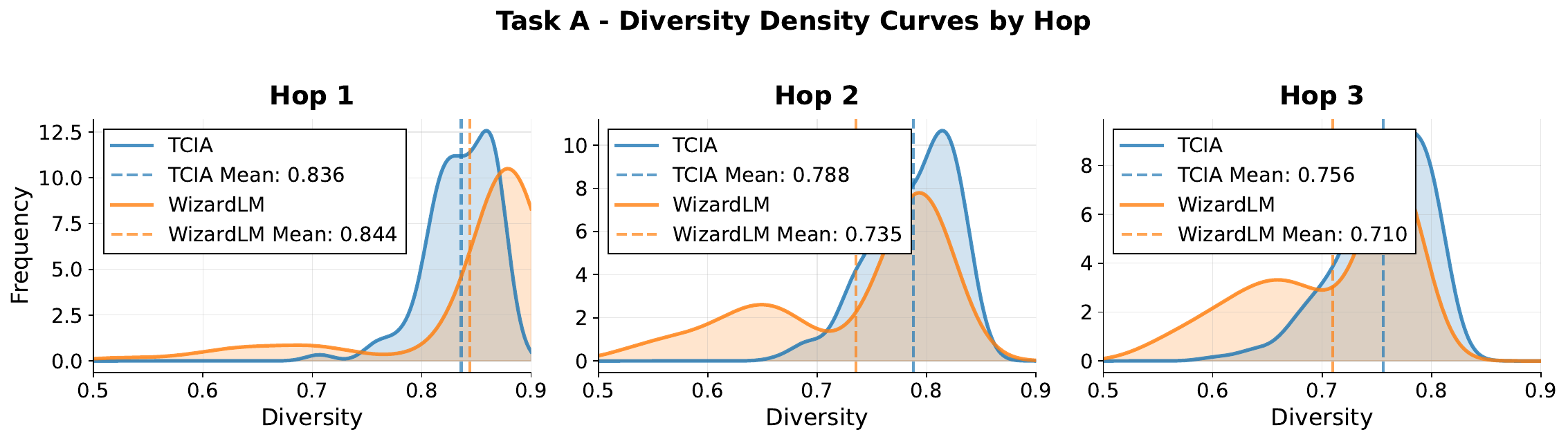} 
\includegraphics[width=0.82\textwidth]{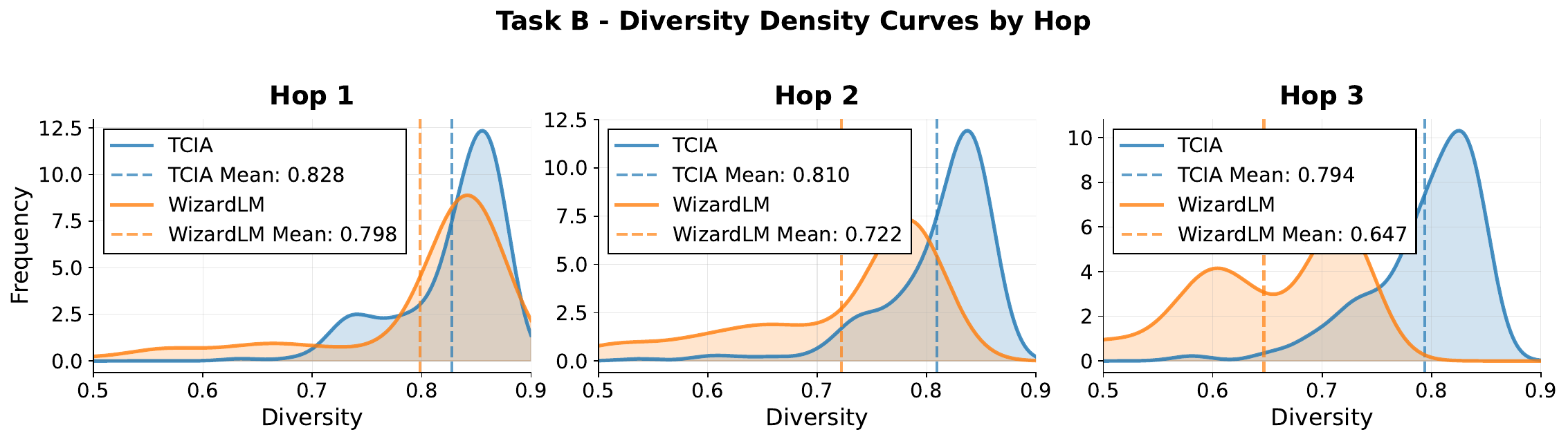} 
\includegraphics[width=0.82\textwidth]{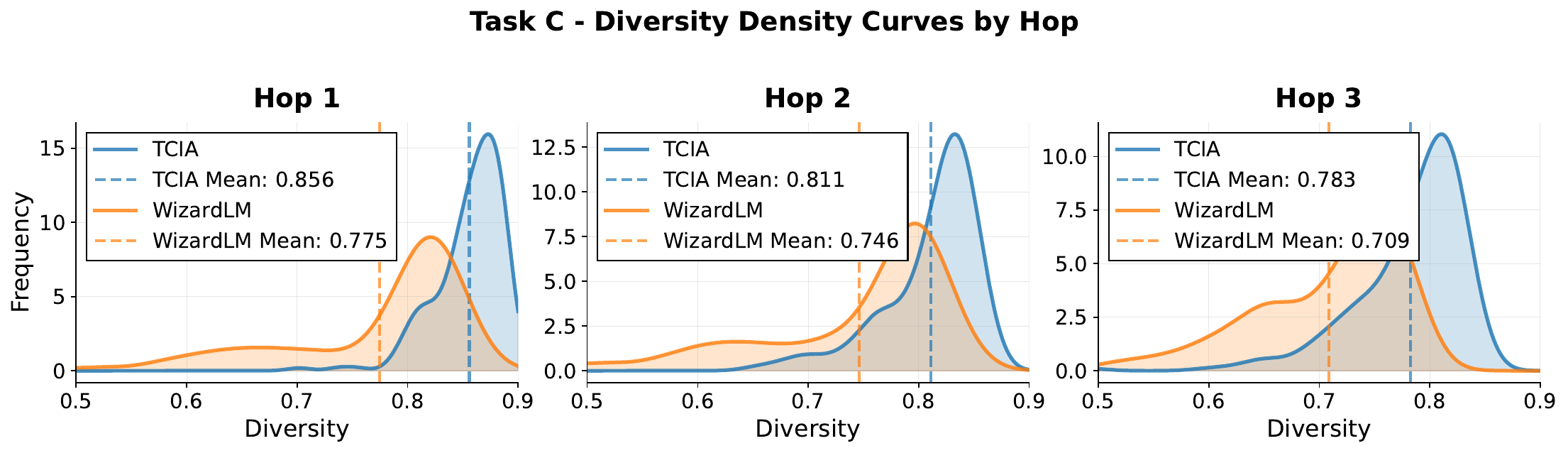} 
\includegraphics[width=0.82\textwidth]{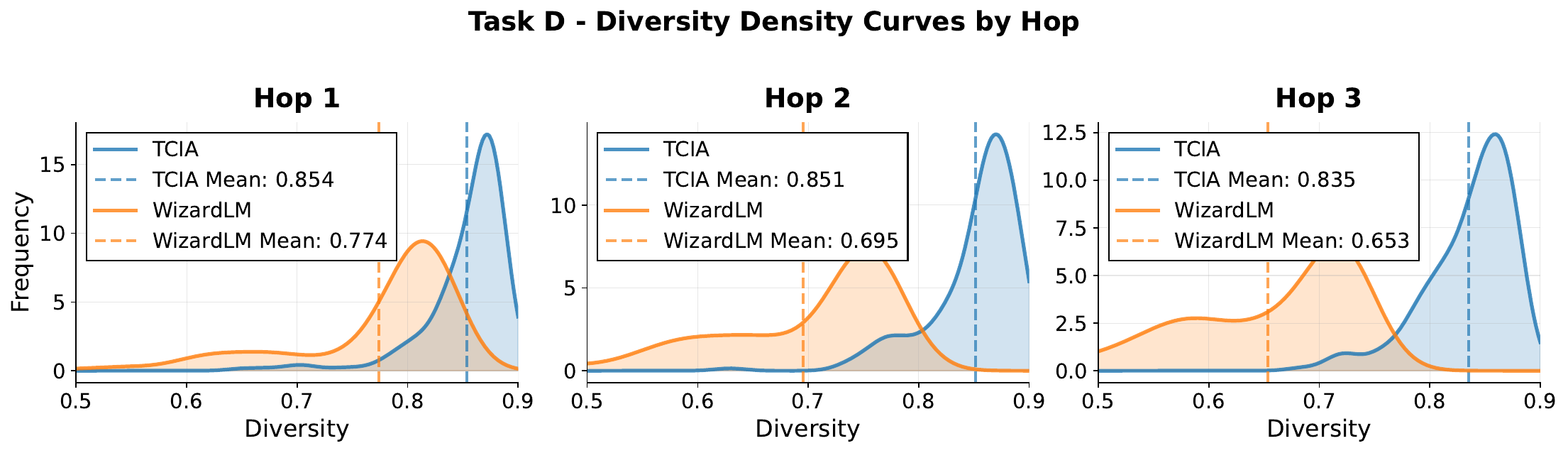} 
\caption{The diversity density plot of TCIA and WizardLM on in-house task A, B, C and D, after 1-3 hops.}
\label{fig: diversity_density_alltask}
\end{figure*}

\clearpage
\begin{figure*}[htbp]
\centering
\includegraphics[width=0.4\textwidth]{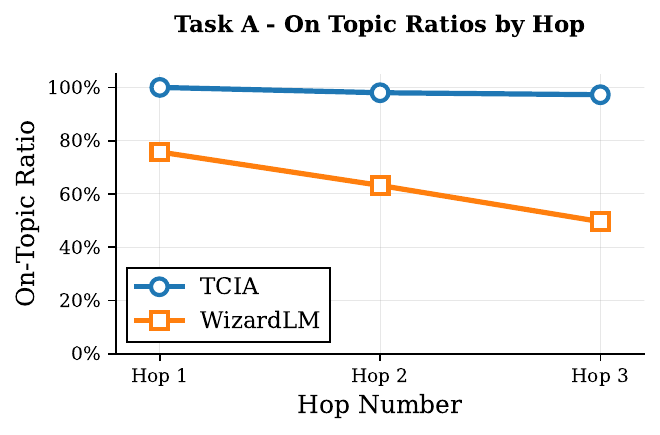}
\includegraphics[width=0.4\textwidth]{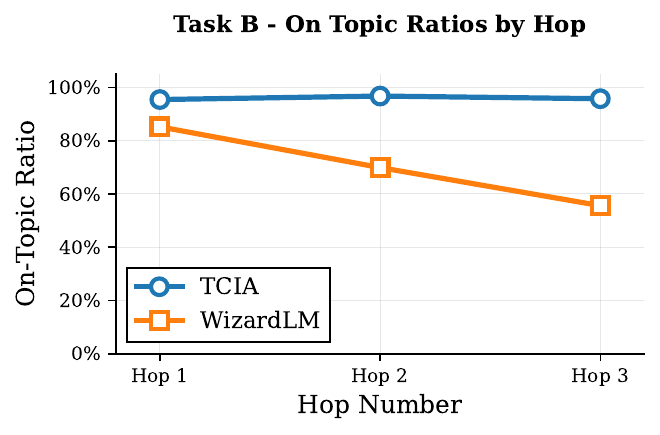}
\includegraphics[width=0.4\textwidth]{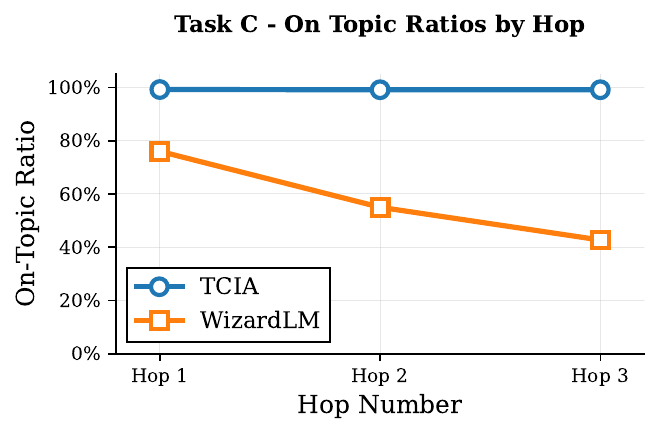} 
\includegraphics[width=0.4\textwidth]{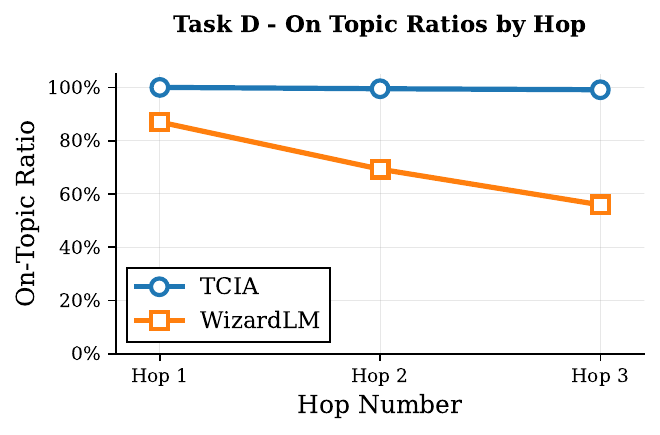} 
\caption{On-task ratio of the generated instructions by TCIA and WizardLM after 1-3 hops, for in-house task A, B, C and D.}
\label{fig: ontask_alltask}
\end{figure*}

\begin{table*}[h]
\centering
\begin{tabular}{|p{0.95\textwidth}|}
\hline
\textbf{Details of TCIA's setup for the Experiment section}
{\scriptsize
\begin{verbatim}
BFS Hyperparameters:
K = 2,700
m = 10
k = 2,000

Instruction Validation LLM: claude-3-5-sonnet-20241024

Output Generation LLMs: 
(1) claude-3-5-sonnet-2024102
(2) claude-3-5-sonnet-20240620
(3) gpt-4o-2024-08-06
(4) gpt-4.1-2025-04-14

Data Quality Filtering LLM: gpt-4.1-2025-04-14
\end{verbatim}
} \\
\hline
\end{tabular}
\caption{Details of TCIA's setup for the Experiment section.}
\label{tab: bfs_hyperparameters}
\end{table*}

\clearpage
\section{Additional Statistical Analysis on the Instruction DB created from Tulu-3 Dataset}
\begin{table*}[h]
\centering
\begin{tabular}{|p{0.95\textwidth}|}
\hline
\textbf{Examples of decomposed instructions and constraint sets of a given user prompt in Tulu-3 Dataset}
{\scriptsize
\begin{verbatim}
{
    "Original User Query": "Create a snippet of Terraform HCL code that create an AWS autoscaling group, and an ALB 
    in front to expose an application to internet.",
    "Complex": "True",
    "Language": "en",
    "Tasks": [
        {
            "Task Type": "Code Generation",
            "Basic Query": "Create code that sets up cloud infrastructure",
            "Constraints": [
                {
                    "category": "Format Constraints",
                    "constraint": "Must be written in Terraform HCL format",
                    "simplified query": "Create code that sets up cloud infrastructure for AWS autoscaling group 
                    and ALB"
                },
                {
                    "category": "Content Constraints",
                    "constraint": "Must include AWS autoscaling group configuration",
                    "simplified query": "Create Terraform code that sets up an ALB to expose an application to 
                    internet"
                },
                {
                    "category": "Content Constraints",
                    "constraint": "Must include AWS Application Load Balancer (ALB) configuration",
                    "simplified query": "Create Terraform code that sets up an autoscaling group"
                },
                {
                    "category": "Content Constraints",
                    "constraint": "Must configure ALB to expose application to internet",
                    "simplified query": "Create Terraform code that sets up an autoscaling group and ALB"
                }
            ]
        }
    ]
}
\end{verbatim}
} \\
\hline
\end{tabular}
\caption{Examples of decomposed instructions and constraint sets of a given user prompt in Tulu-3 Dataset. The decomposition is conducted using the decompose prompt in Table \ref{tab: decompose prompt} via GPT-4.1.}
\label{tab: Tulu3_decompose_example}
\end{table*}

\begin{table*}[h]
\footnotesize
\centering
\begin{tabular}{p{100mm}|p{20mm}|p{30mm}}
\hline
Task Type & Total number of queries & \# unique constraints/\# of unique base query \\ \hline
Mathematical Problem Solving & 144244 & 2.84 \\ \hline
Mathematical Word Problem Solving & 89221 & 3.44 \\ \hline
Mathematical and Engineering Calculations & 51785 & 2.47 \\ \hline
Technical and Creative Problem-Solving Across Multiple Domains & 30030 & 2.52 \\ \hline
Mathematical Optimization and Linear Programming Problem Solving & 28348 & 2.43 \\ \hline
Mathematical Function Creation and Analysis & 28316 & 3.31 \\ \hline
Mathematical Analysis & 27827 & 2.44 \\ \hline
Code Generation and Implementation & 21111 & 2.77 \\ \hline
Translation & 17944 & 3.00 \\ \hline
Mathematical Probability and Statistical Analysis & 17820 & 2.43 \\ \hline
Creative Writing and Composition & 16181 & 4.61 \\ \hline
Information Retrieval & 15144 & 1.82 \\ \hline
Code Implementation Questions & 13677 & 2.76 \\ \hline
Natural Language Inference & 12510 & 14.59 \\ \hline
Multiple Choice Question Answering & 9157 & 1.86 \\ \hline
Mathematical Derivation & 7858 & 2.25\\ \hline
Explanation & 6696 & 2.03\\ \hline
Question Answering & 6309 & 1.26 \\ \hline
Text and Document Generation & 6153 & 2.68 \\ \hline
Information Request & 5644  & 2.02 \\ \hline
Statistical Analysis & 5467  & 2.27 \\ \hline
\end{tabular}
\caption{The top 20 task types of Tulu-3 Dataset, extracted using the decompose prompt in Table \ref{tab: decompose prompt} (shown in the ``Task Type'' field  of the LLM output, e.g. Table \ref{tab: Tulu3_decompose_example}). We use all-mpnet-base-v2 embedding and DBSCAN clustering algorithm to cluster them, resulting in total of 9728 task types.}
\label{tab: Tulu3_top20_tasks}
\end{table*}

\begin{figure*}[h]
\centering
\includegraphics[width=0.4\textwidth]{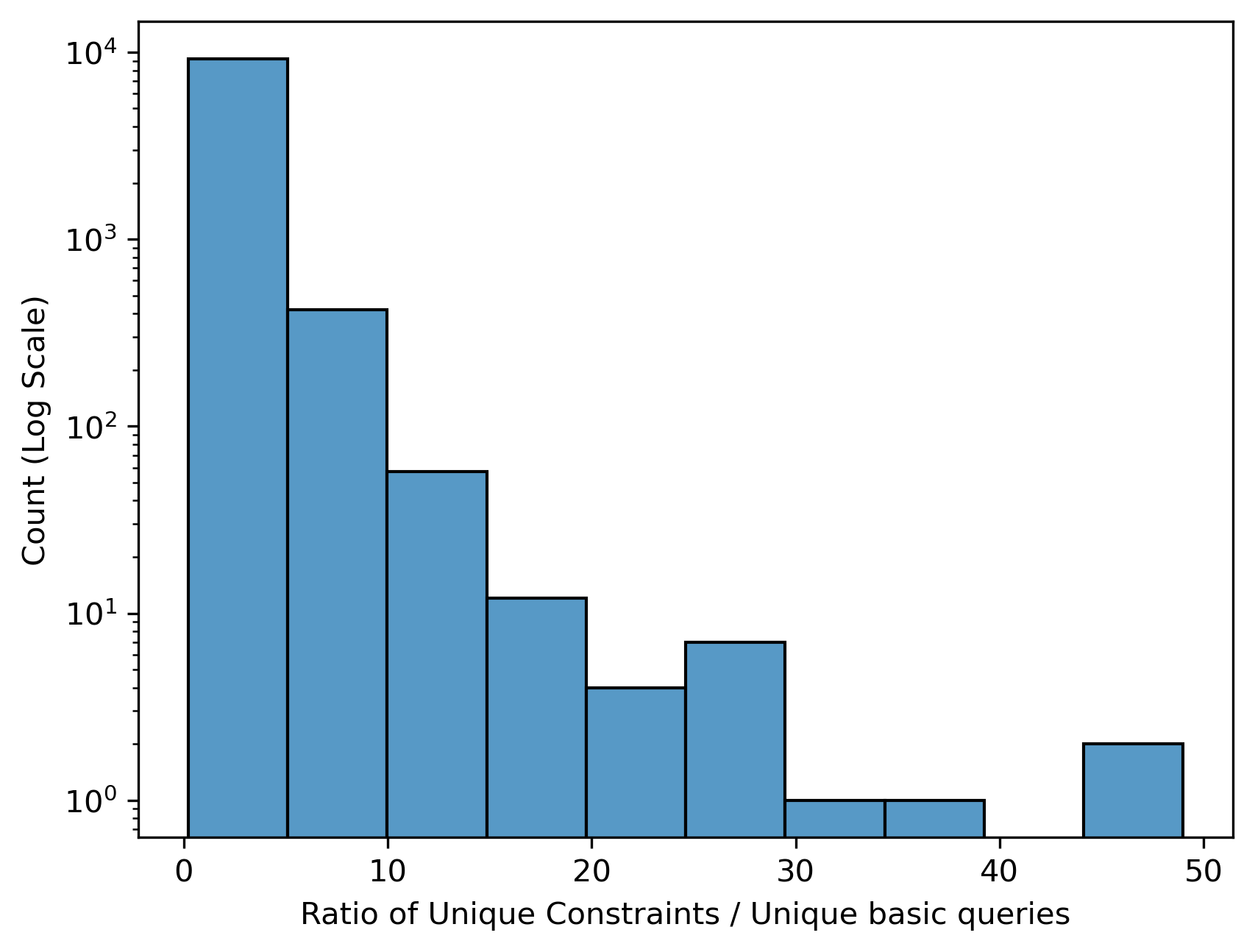}
\includegraphics[width=0.4\textwidth]{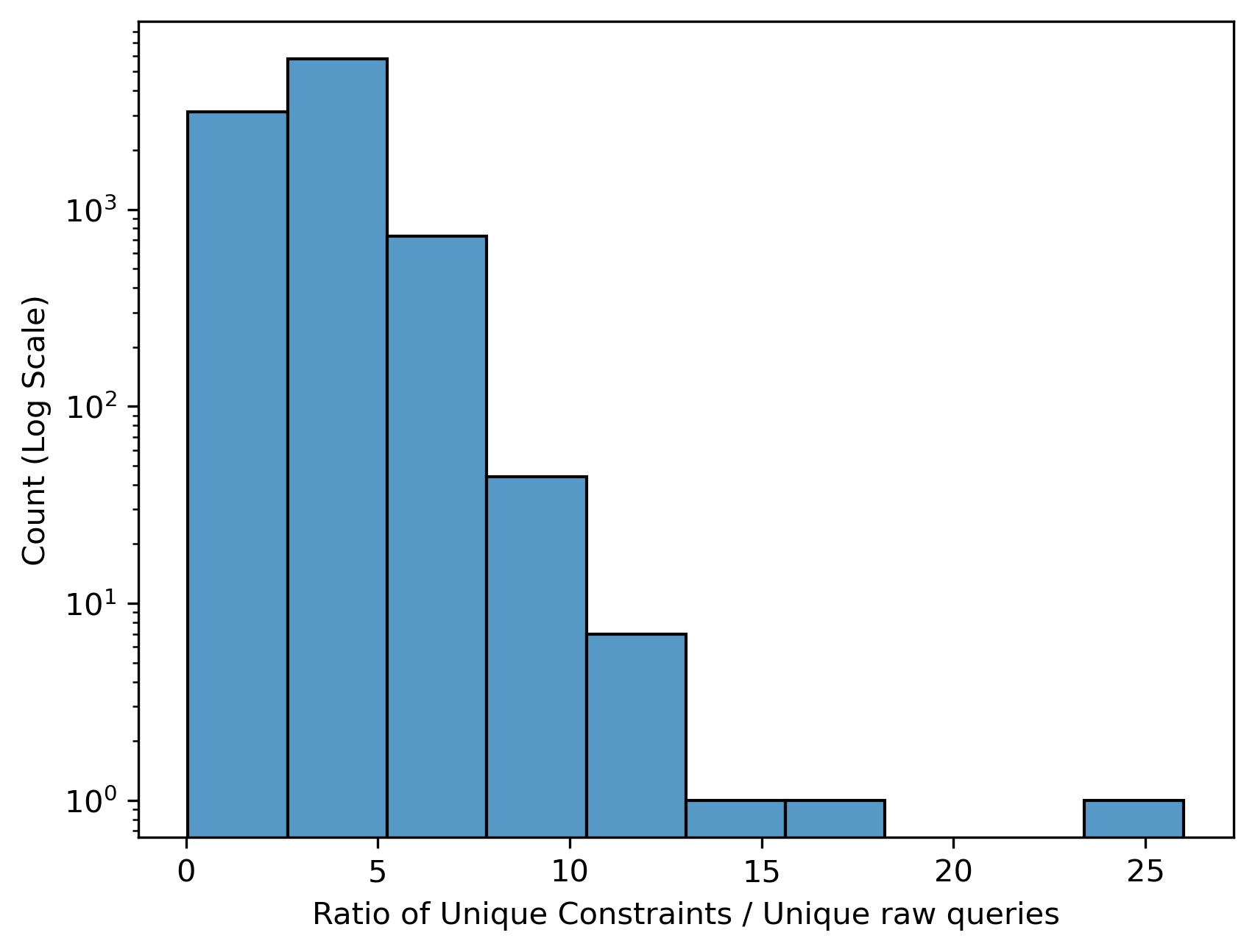}
\caption{Histograms of (1) Left: ratio of unique constraints and unique base queries ratio, (2) Right: ratio of unique constraints and unique raw queries ratio. This shows the distribution of query-constraint ratios in Tulu-3 Dataset. Note that the outliers mainly comes from multiple duplicated queries with different responses, for example ``Write a restaurant review''.}
\label{fig: Tulu3_queryconstraint_ratio}
\end{figure*}

\begin{table*}[h]
\footnotesize
\centering
\begin{tabular}{p{40mm}|p{100mm}|p{20mm}}
\hline
Constraint Type & Example constraints & Occurrences \\ \hline
Probability and Statistical Calculations with Specific Parameters & Must use softmax function $\sigma(z_i) = \exp(z_i)/\sum\exp(z_j)$, The context involves random selection of voters regarding mayor approval, Must consider six specific lines: AB, AC, AD, BC, BD, and CD & 34989 \\ \hline
Poisson Distribution Applications with Various Lambda Parameters &
Must use Poisson distribution with $\lambda$ from previous calculation, Must use Poisson distribution with $\lambda=1200$, Must use Poisson distribution with $\lambda=15$ & 1113 \\\hline
Normal Distribution Calculations with Specified Parameters & 
Must use normal distribution $\mathcal{N}(\mu, \sigma^2)$, Must use normal distribution with mean score of 75 points and standard deviation of 8 points, Must use normal distribution assumption for calculations & 1062 \\\hline
Binomial Distribution Probability Calculations & 
Must use binomial distribution, Must use binomial distribution for probability calculation, Must use binomial distribution for calculation & 301 \\\hline
Time-Based Normal Distribution Parameter Specifications & 
Normal distribution with mean of 10 hours and standard deviation of 2 hours, Must use normal distribution with mean of 2 hours and standard deviation of 0.5 hours, Must use normal distribution with known standard deviation $\sigma=2$ seconds & 281 \\\hline
Advanced Probability Theory and Mathematical Notation Requirements & 
Must use advanced statistical techniques, Must use mathematical notation and probability theory formalism, Must express the solution in mathematical/statistical notation & 170 \\\hline
Independent Events Requirement in Probability Problems & 
Must assume events are independent, All probabilities must be treated as independent events, Probabilities must be treated as independent events & 168 \\\hline
Statistical Confidence Level and Interval Requirements & 
Must achieve at least 90\% confidence level, Must use 95\% confidence interval, 95\% confidence interval required & 122 \\\hline
Poisson Distribution Modeling for Sports Goal Statistics & 
Must use Poisson distribution model for goal-scoring rates, Goals must follow a Poisson distribution with mean of 3 goals per game, Must use Poisson distribution with mean ($\lambda$) of 3.5 goals per game & 113 \\\hline
Use Given Transition Matrix for Markov Chain Calculations & 
Must use specific transition matrix P with given values, Must use the given 3x3 Markov transition matrix P with specific values, Must use a specific 3x3 transition probability matrix with given values & 90 \\\hline
\end{tabular}
\caption{Top 10 occurred example constraints in the task type ``Mathematical Probability and Statistical Analysis'' (the top 10th task in Table \ref{tab: Tulu3_top20_tasks}).}
\label{tab: Tulu3_top10_constraints}
\end{table*}

\begin{figure*}[h]
\centering
\includegraphics[width=\textwidth]{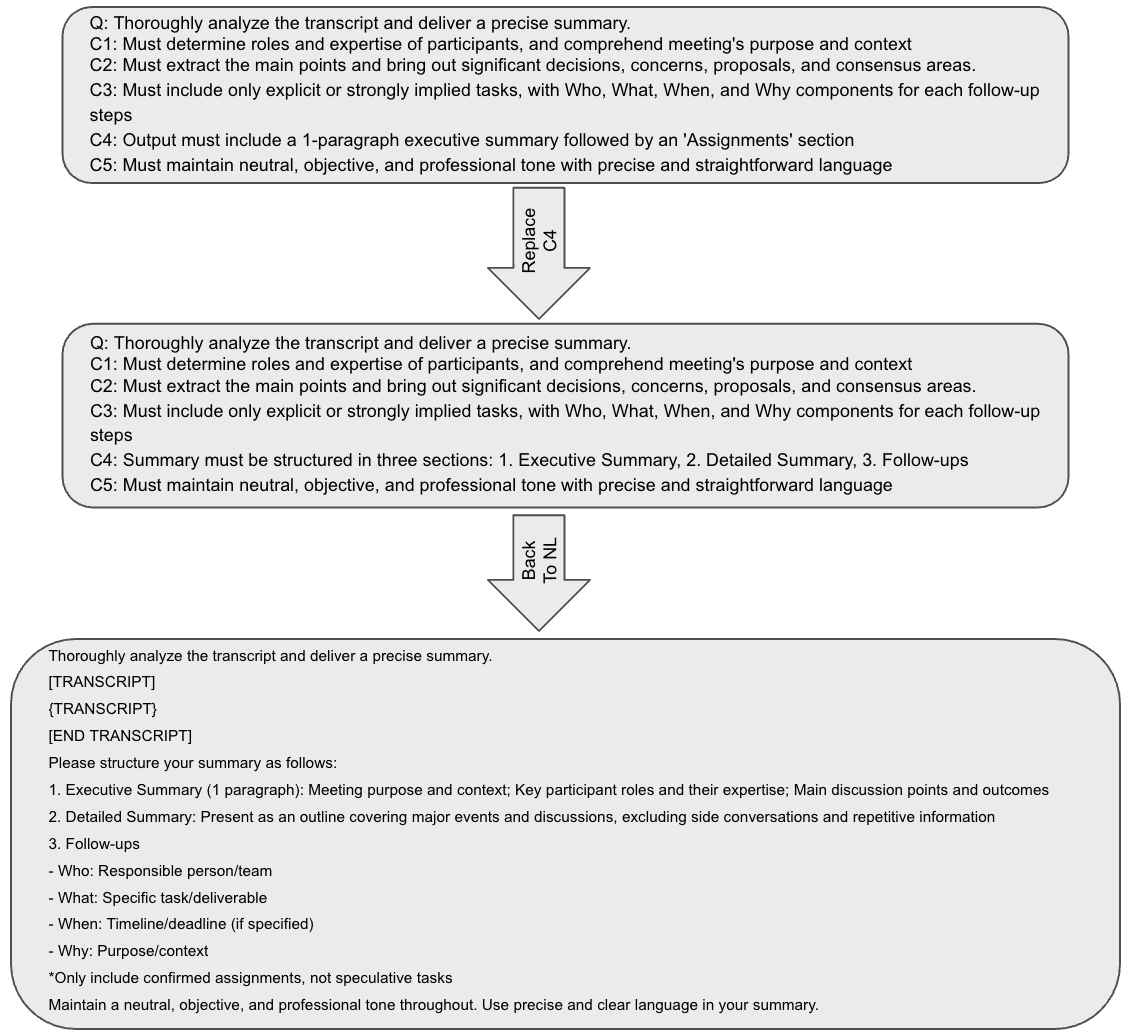}
\caption{Toy example of TCIA on our in-house task B. Here, we use the seed prompt in Table \ref{tab: gpt_meetingsummaryprompt} to first decompose into a base query $Q$ and constraints $\mathbf{C}$. Then the BFS algorithm is applied with only one operation: replace constraint 4. Lastly, $\{Q, \mathbf{C}\}$ is converted back to natural language prompt, which will subsequently go through instruction validation.}
\label{fig: TCIAExample_TaskB}
\end{figure*}
\end{document}